\documentclass{article}

\usepackage{graphicx}
\usepackage{arxiv}

\usepackage[utf8]{inputenc} % allow utf-8 input
\usepackage[T1]{fontenc}    % use 8-bit T1 fonts
\usepackage{hyperref}       % hyperlinks
\usepackage{url}            % simple URL typesetting
\usepackage{booktabs}       % professional-quality tables
\usepackage{amsfonts}       % blackboard math symbols
\usepackage{nicefrac}       % compact symbols for 1/2, etc.
\usepackage{microtype}      % microtypography

\title{From Machine to Machine: An OCT-trained Deep Learning Algorithm for Objective Quantification of Glaucomatous Damage in Fundus Photographs}

\author{Felipe A. Medeiros\thanks{Corresponding author} \\
  Department of Ophthalmology\\
  Duke University\\
  Durham, NC\\
  \texttt{felipe.medeiros@duke.edu} \\
   \And
  Alessandro A. Jammal \\
  Department of Ophthalmology\\
  Duke University\\
  Durham, NC\\
  \And
  Atalie C. Thompson \\
  Department of Ophthalmology \\
  Duke University \\
  Durham, NC \\}

\begin{document}
\maketitle

\begin{abstract}
Previous approaches using deep learning algorithms to classify glaucomatous damage on fundus photographs have been limited by the requirement for human labeling of a reference training set. We propose a new approach using spectral-domain optical coherence tomography (SDOCT) data to train a deep learning algorithm to quantify glaucomatous structural damage on optic disc photographs.
The dataset included 32,820 pairs of optic disc photos and SDOCT retinal nerve fiber layer (RNFL) scans from 2,312 eyes of 1,198 subjects. A deep learning convolutional neural network was trained to assess optic disc photographs and predict SDOCT average RNFL thickness. The performance of the algorithm was evaluated in an independent test sample. The mean prediction of average RNFL thickness from all 6,292 optic disc photos in the test set was 83.3$\pm$14.5 $\mu$m, whereas the mean average RNFL thickness from all corresponding SDOCT scans was 82.5$\pm$16.8 $\mu$m (P = 0.164). There was a very strong correlation between predicted and observed RNFL thickness values (r = 0.832; P<0.001), with mean absolute error of the predictions of 7.39 $\mu$m. The areas under the receiver operating characteristic curves for discriminating glaucoma from healthy eyes with the deep learning predictions and actual SDOCT measurements were 0.944 (95$\%$ CI: 0.912- 0.966) and 0.940 (95$\%$ CI: 0.902 - 0.966), respectively (P = 0.724).
In conclusion, we introduced a novel deep learning approach to assess optic disc photographs and provide quantitative information about the amount of neural damage. This approach could potentially be used to diagnose and stage glaucomatous damage from optic disc photographs. 

\end{abstract}

% keywords can be removed

\section{Introduction}
Glaucoma is a progressive optic neuropathy that results in characteristic changes to the optic disc and retinal nerve fiber layer.[1] Although damage from glaucoma is irreversible, early treatment can usually prevent or slow down progression to functional damage and visual impairment.[2]

Assessment of structural damage is essential for early detection of glaucoma. Fundus photographs are a low-cost and easy to perform method to document and identify optic disc features characteristic of glaucoma. However, it is well-established that subjective evaluation of optic disc photographs suffers from low reproducibility, even when performed by expert graders.[3,4,5] In addition, graders frequently under- or over-estimate glaucoma likelihood when evaluating disc photographs.[6] Recent progress in artificial intelligence and machine learning has led to the development of algorithms capable of objective assessment of fundus photographs for identification of signs of ocular diseases.[7,8,9] Li et al[7] evaluated the ability of a deep learning neural network algorithm to identify signs of glaucomatous neuropathy on color fundus photographs. The authors reported excellent sensitivity and specify to diagnose "referable" glaucomatous optic neuropathy, which was defined based on subjective grading of the photographs by a group of trained ophthalmologists.

A fundamental step in the development of any machine learning algorithm is the training process by which the algorithm "learns" to correctly make classifications and predictions. Essentially, the algorithm cannot perform better than the reference standard used to train it, and its best hope is to perfectly replicate the classifications or predictions that would have been made by the reference standard. Although the work by Li et al[7] provides important insights into how machine learning could be used to assess glaucomatous damage on fundus photographs, their algorithm suffers from the limitation that subjective grading was used as the reference standard to train the deep learning network.

In recent decades, spectral-domain optical coherence tomography (SDOCT) has become the de facto standard in objective quantification of structural damage in glaucoma.[10]  Measurements of the retinal nerve fiber layer (RNFL) thickness with SDOCT show high reproducibility and have been shown to accurately diagnose the disease, detect its progression, and measure rates of change.[10-13] As such, it is conceivable that a superior machine learning classifier could be obtained if it is trained to evaluate fundus photographs using SDOCT measurements as the reference standard, rather than subjective assessment by graders. Such an algorithm could also be trained to obtain quantitative rather than just qualitative assessments of the amount of neural damage from disc photographs.

In the present work, we report on a novel deep learning algorithm trained to assess optic disc photographs from results of SDOCT and investigate its ability to provide objective quantification of glaucomatous neural loss.

\section{Methods}
\label{sec:headings}
The dataset for this study was collected from the Duke Glaucoma Repository, a database of electronic medical and research records at the Vision, Imaging and Performance (VIP) Laboratory from the Duke Eye Center. The institutional review board from Duke University approved this study, and a waiver of informed consent was provided due to the retrospective nature of this work. All methods adhered to the tenets of the Declaration of Helsinki for research involving human subjects and the study was conducted in accordance with regulations of the Health Insurance Portability and Accountability Act.

The database contained information on comprehensive ophthalmologic examinations during follow-up, diagnoses, medical history, visual acuity, slit-lamp biomicroscopy, intraocular pressure measurements, results of gonioscopy and dilated slit-lamp funduscopic examinations. In addition, the repository contained stereoscopic optic disc photographs (Nidek 3DX, Nidek, Japan), standard automated perimetry (SAP; Humphrey Field Analyzer II, Carl Zeiss Meditec, Inc., Dublin, CA) and Spectralis SOCT (Software version 5.4.7.0, Heidelberg Engineering, GmbH, Dossenheim, Germany) images and data. SAP was acquired with the 24-2 Swedish interactive threshold algorithm (Carl Zeiss Meditec, Inc., Dublin, CA). Only subjects with open angles on gonioscopy were included. Visual fields were excluded if they had more than 33$\%$ fixation losses or more than 15$\%$ false-positive errors. Patients were excluded if they had a history of other ocular or systemic diseases that could affect the optic nerve or the visual field. 
Diagnosis of glaucoma was defined based on the presence of glaucomatous repeatable visual field loss in SAP (pattern standard deviation [PSD] < 5$\%$ or glaucoma hemifield test outside normal limits) and signs of glaucomatous neuropathy as based on records of slit-lamp fundus examination. Glaucoma suspects were those with history of elevated intraocular pressure, suspicious appearance of the optic disc on slit-lamp fundus examination or with other risk factors for the disease. Normal subjects were required to have a normal optic disc appearance on slit-lamp fundus examination in both eyes as well as no history of elevated intraocular pressure and normal SAP results.

Images were acquired with the Spectralis SD-OCT to assess the RNFL. The device uses a dual-beam SDOCT and a confocal laser-scanning ophthalmoscope that employs a super luminescent diode light with a center wavelength of 870 nm and an infrared scan to provide simultaneous images of ocular microstructures. The Spectralis RNFL circle scan was used for this study. A total of 1536 A-scan points were acquired from a 3.45-mm circle centered on the optic disc. Axial length and corneal curvature measurements were entered into the instrument software to ensure accurate scaling of all measurements, and the device's eye tracking capability was used during image acquisition to adjust for eye movements and to ensure that the same location of the retina was scanned over time. Images were manually reviewed to ensure quality, scan centration and no coexistent retinal pathologic abnormalities. Images that had been inverted or clipped, or with signal strength below 15 dB, were excluded. The average circumpapillary RNFL thickness corresponds to the 360 degrees measure automatically calculated by the SDOCT software. 

For each eye of each subject, we considered all the available optic disc photographs that had been acquired over time and matched them to the closest Spectralis SDOCT RNFL scan acquired within 6 months from the photo date.  As subjects were followed over time, multiple pairs of SDOCT and disc photos were available for each subject. This was important in order to increase the heterogeneity of the dataset for deep learning training.

\subsection{Development of the Deep Learning Algorithm}
A deep learning algorithm was trained to predict SDOCT average RNFL thickness from assessment of optic disc photographs. The target value, i.e., the variable we wanted to predict from analysis of optic disc photographs was the SDOCT average RNFL thickness. Therefore, for training the neural network, a pair of train-target consisted of the optic disc photograph and the SDOCT average RNFL thickness value.  The sample of pairs of photos-OCT was split into a training plus validation set (80$\%$) and test sample (20$\%$). Importantly, in order to prevent leakage and biased estimates of test performance, the random sampling process was at the patient level, so no data of any patient was present in both the training and the test samples.

The optic disc stereophotographs were initially preprocessed to derive data for the deep learning algorithm. Each stereoscopic photograph was split creating a pair of photos from the stereo views. The images were then downsampled to 256 x 256 pixels and pixel values were scaled to range from 0 to 1.  Data augmentation was performed to increase heterogeneity of the photographs, reducing the possibility of overfitting and allowing the algorithm to learn the most relevant features. Data augmentation included the following: random lighting, consisting of subtle changes in image balance and contrast; random rotation, consisting of rotations of up to 10 degrees in the image; and random flips, consisting of flipping the image vertically or horizontally. 

We used the Residual deep neural Network (ResNet34) architecture. The ResNet is a deep residual network that allows relatively rapid training of very deep convolutional neural networks in a way that has not been previously possible.[14] In brief, these networks use identity shortcut connections that skip one or more layers and greatly decrease the vanishing gradient problem when training deep networks. In the present work, a ResNet that had been previously trained on the ImageNet dataset was used.[15] However, as the recognition task of the present work largely differs from that of ImageNet, further training was performed by initially unfreezing the last 2 layers. Subsequently, all layers were unfrozen, and training was performed using differential learning rates. The network was trained with minibatch gradient descent of size 64 and Adam optimizer. The best learning rate was found using the cyclical learning method with stochastic gradient descents with restarts. 

As a variant of the training process described above, we also trained the deep learning network to classify optic disc photographs in normal versus abnormal according to the SDOCT average RNFL thickness categorical classification as provided by the instrument's normative database. This was done in order to allow investigation of how deep learning assessment of photographs would perform in classifying and categorizing the presence of damage. For this purpose, a borderline classification of SDOCT average thickness was included as within normal limits (to retain high specificity), generating a binary target variable. The deep learning model then calculated the probability of abnormality based on assessment of optic disc photos. We built heatmaps corresponding to the Gradient-weighted class activation maps over the input images.[16] These heatmaps indicate how important each location of the image is with respect to the class under consideration. This technique allows one to visualize the parts of the image that are most important in the deep neural network classification.

\subsection{Statistical Analysis}
The performance of the deep learning algorithm in quantifying glaucomatous damage in optic disc photographs was evaluated in the test sample by comparing the predictions with the actual SDOCT average RNFL thickness. Generalized estimating equations (GEE) were used to account for the fact that multiple measures were obtained per patient. We calculated the mean absolute error  (MAE) of the predictions as well as Pearson's correlation coefficient and agreement by the Bland-Altman plot 95$\%$ confidence limits of agreement. We also investigated the correspondence between classifications performed by the deep learning system and those given by the SDOCT normative database. 

We also investigated the relationship between predicted and observed values of RNFL thickness and SAP mean deviation (MD) with locally weighted weighted scatterplot smoothing (lowess). Receiver operating characteristic curves were used to assess and compare the ability of the deep learning algorithm on photographs versus SDOCT average RNFL thickness in discriminating eyes with glaucomatous visual field loss from healthy eyes. The ROC curve provides the tradeoff between the sensitivity and specificity. The area under the ROC curve (AUC) was used to summarize the diagnostic accuracy of each parameter. An AUC of 1.0 represents perfect discrimination, whereas an area of 0.5 represents chance discrimination. Sensitivity at fixed specificities of 80$\%$ and 95$\%$ were also reported. To account for using multiple images of both eyes of the same participant in the analyses, a bootstrap resampling procedure was used to derive confidence intervals and P-values, where the cluster of data for the participant was considered as the unit of resampling to adjust standard errors. This procedure has been previously used to adjust for the presence of multiple correlated measurements from the same unit.[17]

\section{Results}

The dataset included 32,820 pairs of optic disc photos and SDOCT scans from 2,312 eyes of 1,198 subjects. The test sample consisted of 6,292 pairs of disc photos and SDOCTs from 463 eyes of 240 subjects. Tables 1 and 2 show demographic and clinical characteristics of the subjects and eyes in the training and test samples, respectively. 

\begin{table}
\setlength{\tabcolsep}{6pt}
 \caption{Demographic and clinical characteristics of the eyes and subjects included in the training sample.}
  \centering
  \begin{tabular}{lccc} 
    \toprule
    \multicolumn{4}{c}{26,528 pairs of disc photos and SDOCT scans from 1849 eyes of 958 subjects}\\
    \cmidrule(r){1-4}
    Parameter & Normal & Suspect & Glaucoma\\
    \midrule
   Number of eyes & 476 & 674 & 699 \\
   Number of images & 3,982 & 13,410 & 9,136 \\
   Age (years) & 57.8$\pm$13.9 & 65.4$\pm$11.1 & 69.7$\pm$11.2 \\
   Gender, $\%$Female &	64.7 & 60.5 & 53.1 \\
   Race ($\%$) \\
   Caucasian & 56.7 & 61.8 & 60.2\\
   African-American & 43.3 & 38.2 & 39.8 \\	
   SAP MD (dB) &	0.05$\pm$1.10 & -0.62$\pm$1.91 & -7.37$\pm$6.95 \\
   SAP PSD (dB) &	1.60$\pm$0.40 & 1.94$\pm$1.10 &	6.40$\pm$3.94 \\
   SDOCT RNFL Thickness ($\mu$m)&96.8$\pm$10.9 & 89.1$\pm$12.8 & 68.3$\pm$14.8 \\
    \bottomrule
  \end{tabular}
  \label{tab:table}
\end{table}

\begin{table}
\setlength{\tabcolsep}{6pt}
 \caption{Demographic and clinical characteristics of the eyes and subjects included in the test sample.}
  \centering
  \begin{tabular}{lccc} 
    \toprule
   \multicolumn{4}{c}{ 6,292 pairs of disc photos and SDOCT scans from 463 eyes of 240 subjects}\\
    \cmidrule(r){1-4}
    Parameter & Normal & Suspect & Glaucoma\\
    \midrule
   Number of eyes & 128	& 164 &	171 \\
   Number of images & 877 &	3,345 &	2,070 \\
   Age (years) & 56.5$\pm$15.9 & 65.5$\pm$11.3 & 68.1$\pm$12.8 \\
   Gender, $\%$Female &	59.1 & 64.5 & 45.2 \\
   Race ($\%$) \\
   Caucasian & 51.8 & 65.8 & 58.0 \\
   African-American & 48.2 & 34.2 & 42.0 \\	
   SAP MD (dB) &	-0.06$\pm$1.10 & -0.62$\pm$2.36 & -7.65$\pm$6.90 \\
   SAP PSD (dB) &	1.61$\pm$0.35 & 2.00$\pm$1.19 &	6.63$\pm$3.99 \\
   SDOCT RNFL Thickness ($\mu$m) & 97.6$\pm$9.3 & 87.1$\pm$12.5 & 68.8$\pm$16.0 \\
  \bottomrule
  \end{tabular}
  \label{tab:table}
\end{table}

Figure 1 shows the relationship between deep learning predictions of average RNFL thickness from optic disc photographs and the actual SDOCT measurements in the test sample. The mean prediction of average RNFL thickness from all 6,292 optic disc photos was 83.3$\pm$14.5 $\mu$m, whereas the mean average RNFL thickness from all the 6,292 corresponding SDOCT scans was 82.5$\pm$16.8  m (P = 0.164; GEE).  There was a very strong correlation between the predicted and the observed RNFL thickness values (Pearson's r = 0.832; R\textsuperscript{2} = 69.3$\%$; P<0.001), with MAE of 7.39 $\mu$m. The 95$\%$ confidence limits of agreement ranged from -18.5$\mu$m to 17.5$\mu$m. Figure 2 shows violin plots illustrating the distribution of predicted and observed RNFL thickness values in normal, suspect, and glaucomatous eyes in the test sample.  Average predictions were 96.1$\pm$7.8 $\mu$m, 87.5$\pm$9.9 $\mu$m and 71.0$\pm$14.4 $\mu$m in normal, suspect and glaucomatous eyes, respectively. There was a statistically significant difference between average predictions for all pairwise comparisons between the 3 groups (P<0.001, GEE). Corresponding numbers for SDOCT mean average RNFL thickness in the three groups were 97.6$\pm$9.3 $\mu$m, 87.1$\pm$12.5 $\mu$m and 68.8$\pm$16.0 $\mu$m (P<0.001 for all pairwise comparisons, GEE).

\begin{figure}[h!]
  \caption{Scatterplot and histograms illustrating the relationship between predictions obtained by the deep learning algorithm evaluating optic disc photographs and actual average retinal nerve fiber layer thickness measurements from spectral-domain optical coherence tomography (OCT). Data is from the independent test set.}
  \centering
  \includegraphics[width=0.5\textwidth]{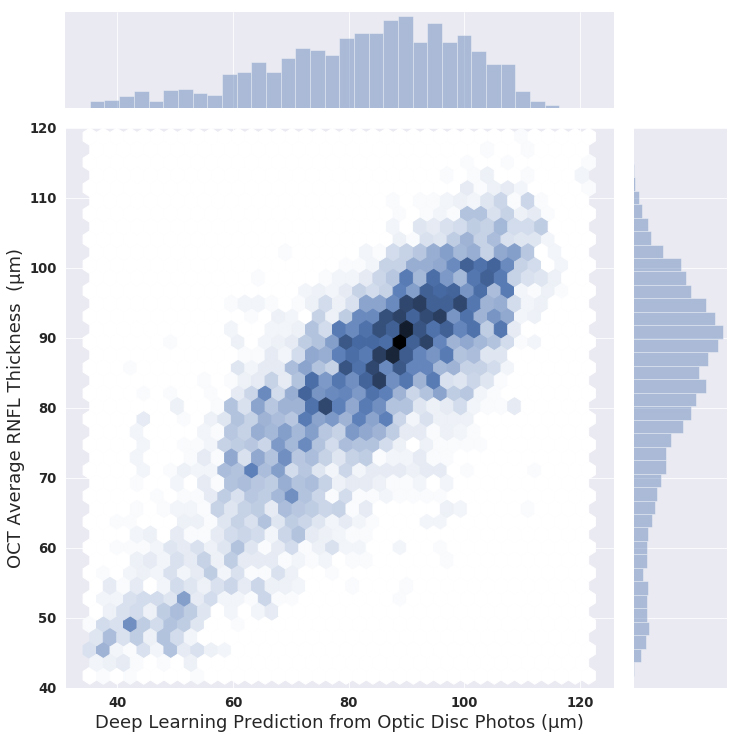}
\end{figure}

\begin{figure}[h!]
  \caption{Violin plots illustrating the distribution of deep learning predictions and optical coherence tomography average retinal nerve fiber layer thickness in normal, suspect and glaucomatous eyes.}
  \centering
  \includegraphics[width=0.60\textwidth]{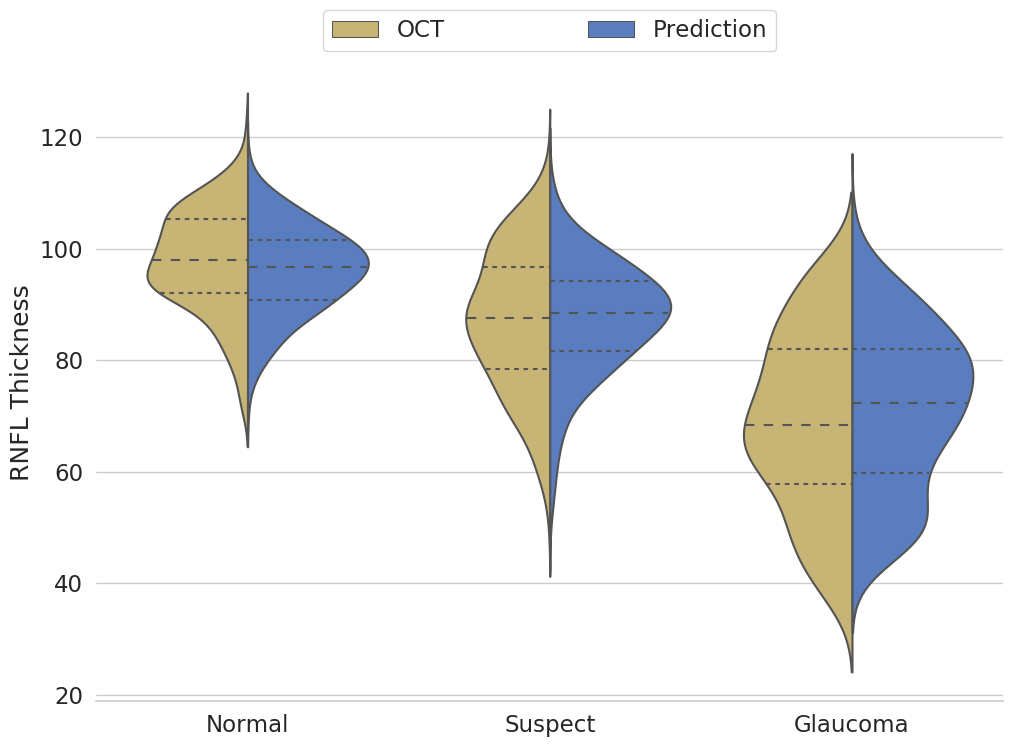}
\end{figure}

There was a significant correlation between RNFL thickness predicted from photos and SAP MD (Pearson's r = 0.61; R\textsuperscript{2} = 37$\%$), which was similar to that between actual SDOCT RNFL thickness and MD (Pearson's r = 0.59; R\textsuperscript{2} = 35$\%$). Figure 3 illustrates the relationship between SAP MD versus observed and predicted RNFL thickness values. The ROC curve area for discriminating glaucomatous from normal eyes with the deep learning optic disc photo predictions was 0.944 (95$\%$ CI: 0.912- 0.966), whereas the ROC curve area for actual SDOCT average RNFL thickness was 0.940 (95$\%$ CI: 0.902 - 0.966). There was no statistically significant difference between the ROC curve areas (P = 0.724). For specificity at 95$\%$, the predicted measurements had sensitivity of 76$\%$ (95$\%$ CI: 64$\%$ - 84$\%$), whereas actual SDOCT measurements had sensitivity of 73$\%$ (95$\%$ CI: 59$\%$ - 85$\%$). For specificity at 80$\%$, the predicted measurements had sensitivity of 90$\%$ (95$\%$ CI: 82$\%$ - 95$\%$), whereas actual SDOCT measurements had sensitivity of 90$\%$ (95$\%$ CI: 83$\%$ - 95$\%$). 

When compared to SDOCT classifications provided by the instrument's normative database, the deep learning network achieved an accuracy of 83.7$\%$. Figure 4 shows examples of optic disc photos and corresponding class activation maps (heatmaps) of the deep learning network. The heatmaps show that the activations were most strongly found in the area of the optic nerve and adjacent retinal nerve fiber layer on the photographs, indicating that these areas were the most important for the network classifications. Figures 5 and 5 show several random examples of optic disc photos from the test sample where the deep learning algorithm correctly and incorrectly predicted the classification given by the SDOCT average RNFL thickness, respectively. 

\begin{figure}
  \caption{Scatterplots with fitted locally weighted scatterplot smoothing (LOWESS) curves illustrating the relationship between visual field mean deviation and average retinal nerve fiber layer thickness from deep learning optic disc photographs predictions (right) and actual optical coherence tomography (left).}
  \centering
  \includegraphics[width=0.75\textwidth]{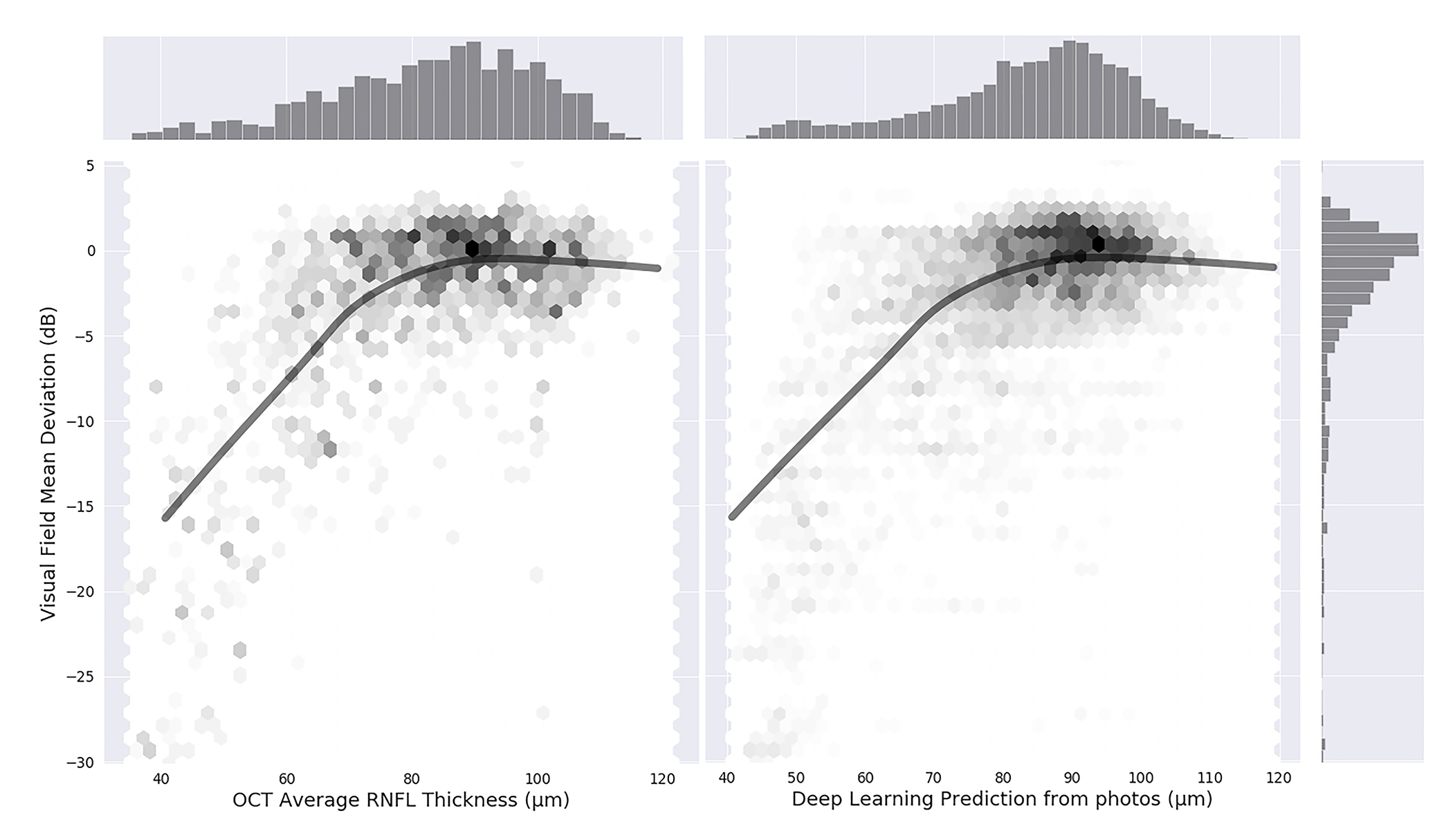}
\end{figure}

\begin{figure}
  \caption{Class activation maps (heatmaps) showing the regions of the photograph that had greatest weight in the deep learning algorithm classification. A is from a normal eye, B from a glaucoma suspect, and C and D are from glaucomatous eyes.}
  \centering
  \includegraphics[width=0.75\textwidth]{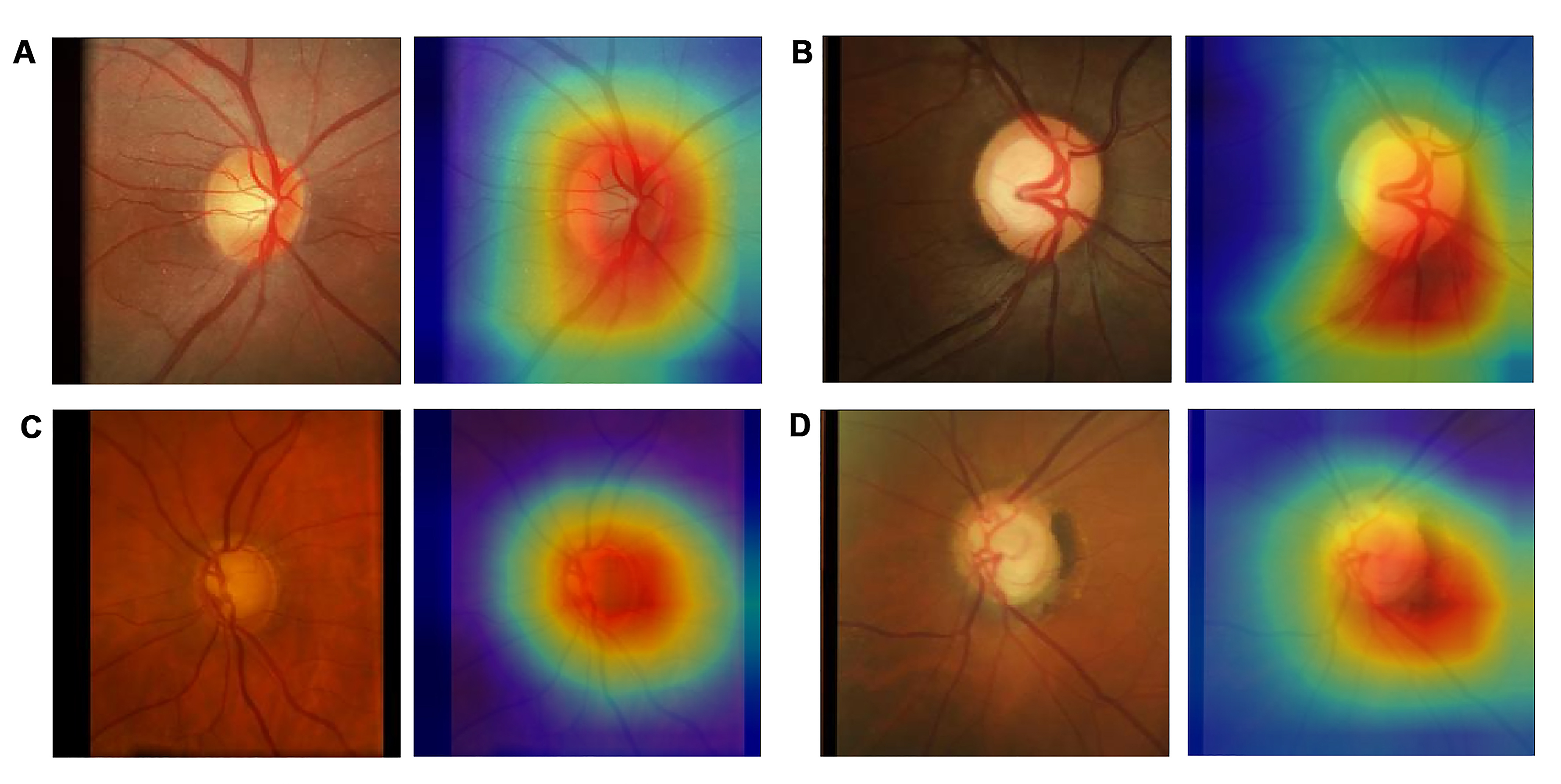}
\end{figure}

\begin{figure}[h!]
  \caption{Random examples of optic disc photographs that were correctly classified according to the reference classification of the Spectralis spectral domain-optical coherence tomography (SDOCT) normative database for average retinal nerve fiber layer thickness (RNFL). Above each photo is shown the SDOCT average thickness measurement, the deep learning (DL) prediction of average RNFL thickness from the optic disc photograph, and the probability of abnormality estimated by the DL algorithm.}
  \centering
  \includegraphics[width=0.8\textwidth]{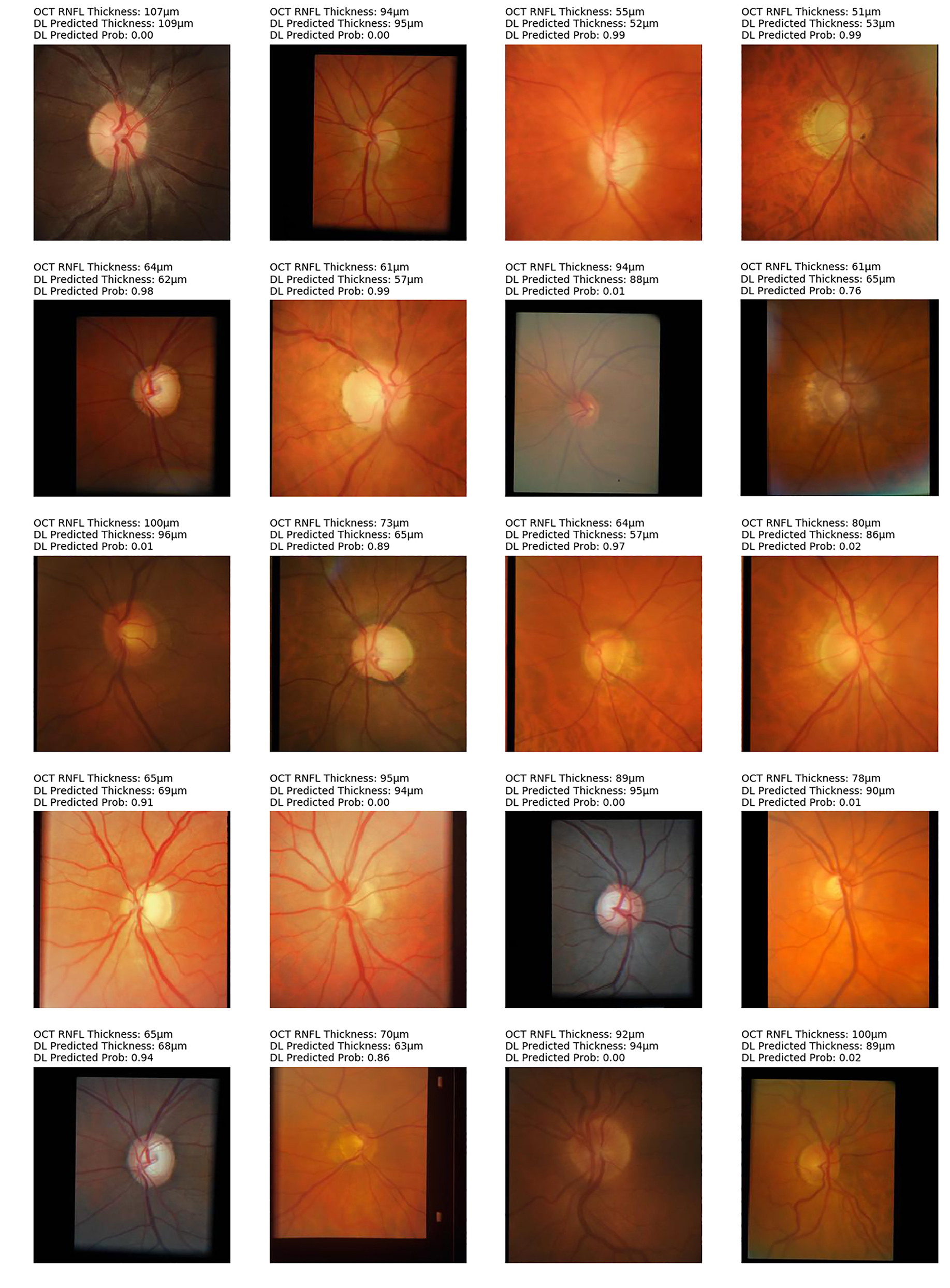}
\end{figure}

\begin{figure}[h!]
  \caption{Random examples of optic disc photographs that were incorrectly classified according to the reference classification of the Spectralis spectral-domain optical coherence tomography (SDOCT) normative database for average retinal nerve fiber layer thickness. Above each photo is shown the SDOCT average thickness measurement, the deep learning (DL) prediction of average RNFL thickness from the optic disc photograph, and the probability of abnormality estimated by the DL algorithm.}
  \centering
  \includegraphics[width=0.8\textwidth]{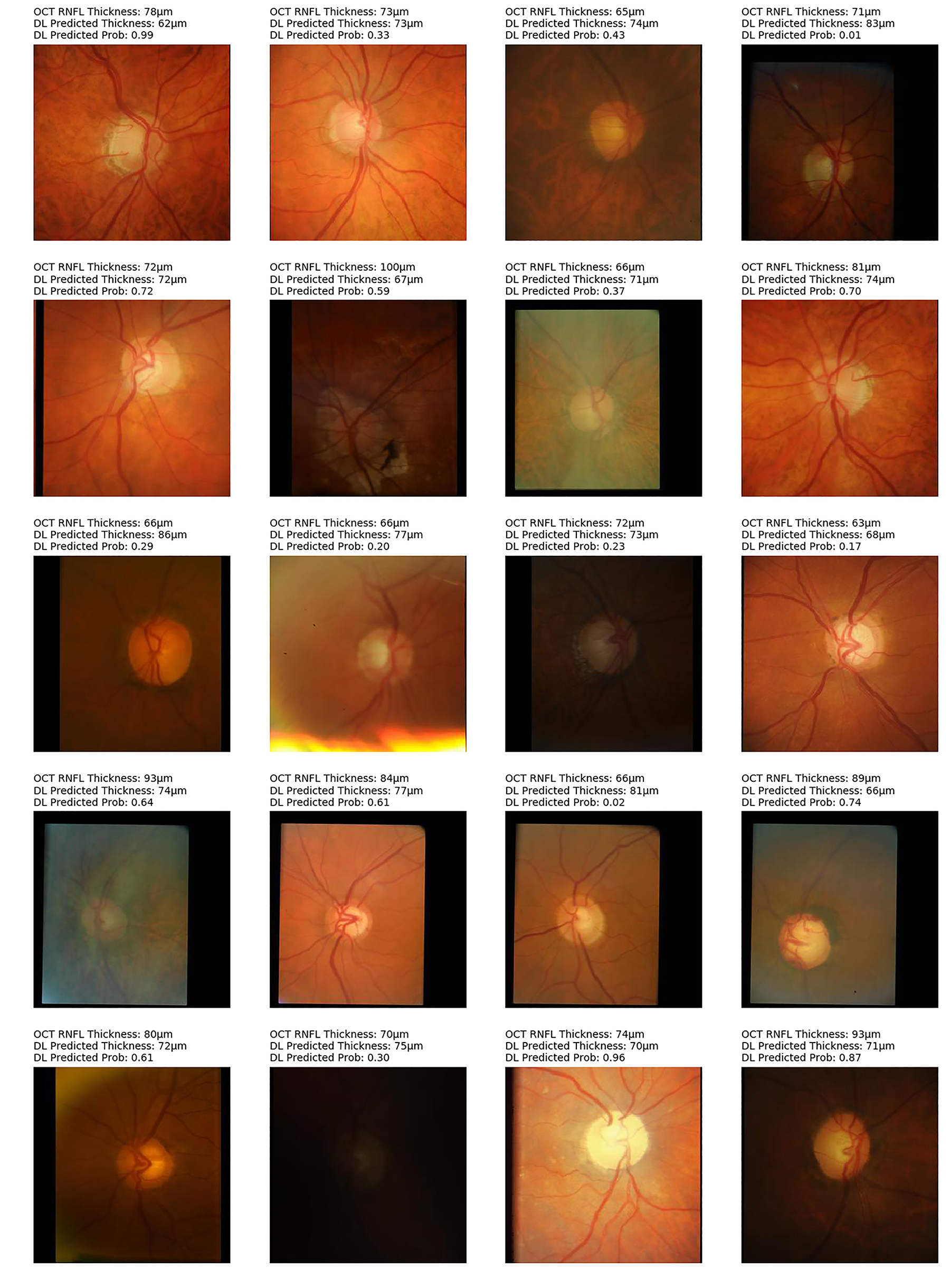}
\end{figure}

\section{Discussion}

In the present study, we developed and validated a novel deep learning algorithm to assess optic disc photographs for the presence of glaucomatous damage. In contrast to previous works in this area, our algorithm was capable of outputting continuous predictions of estimated RNFL thickness, therefore allowing for quantitative assessment of the amount of neural damage on disc photos. This was achieved by training the network with RNFL thickness measurements extracted from SDOCT. To the best of our knowledge, such an approach has not been previously described in the literature. 

Previous investigators have described machine learning approaches to assess optic disc photographs for glaucomatous damage.[7,8] In these investigations, human graders were asked to label the photos for the presence of glaucomatous damage and such labeling was used as the reference standard to train the classifier. Such an approach presents several limitations. Gradings of optic disc photos by human graders are subjective and known to have relatively poor reproducibility.[3,4,5] Furthermore, misclassifications are very likely to occur. For example, graders tend to frequently misclassify eyes with physiologic large cups as having glaucoma and they often miss signs of glaucomatous damage in eyes with small optic discs. If a machine classifier is trained using human labeling of optic disc photos, it will essentially replicate those errors and is likely to have poor performance as a screening tool, even if it shows high accuracy when compared to the human gradings. Our study proposes a novel approach in training the classifier by using RNFL thickness measurements extracted from SDOCT. This presents some obvious advantages. First, it provides an objective and reproducible metric to serve as a target. Average RNFL thickness measurements have been shown to accurately detect glaucomatous damage and can, for example, discriminate eyes with large physiologic cups from those with actual glaucomatous damage.[18,19] In addition, training a network with objective SDOCT data obviates the need for the time-consuming task of subjective labeling by human graders. 

The predictions obtained from deep learning analysis of optic disc photographs showed very strong correlation with the actual RNFL thickness measurements in the independent test sample. Furthermore, the MAE of the predictions was only approximately 7 $\mu$m. Interestingly, a previous study[20] reported an R\textsuperscript{2} of 70$\%$ for the correlation between two different SDOCT devices, and another study reported Bland-Altman limits of agreement that are not far from those reported in our investigation.[21]  As a result, it was not surprising that the predictions performed well to discriminate eyes with glaucomatous visual field loss from healthy eyes. In fact, the ROC curve area for the predictions was almost identical to that for actual SDOCT RNFL thickness values. This provides important validation of our deep learning model. As described before, previous models have essentially attempted to replicate human grading of photographs but have provided no evidence that the algorithm classifications or predictions actually corresponded to clinically relevant outcomes. By showing a correspondence to visual field loss in a similar degree to SDOCT, our work provides essential validation of the quantitative deep learning approach to assess disc photographs.

Although SDOCT has become the reference standard for quantification of structural damage in glaucoma, assessment of optic disc photographs may present several advantages. SDOCTs are still generally expensive and non-portable machines which can be difficult to implement in screening settings, notably in underserved populations. In contrast, photographs may provide a quick and inexpensive method for documenting the optic disc appearance. Recent work has demonstrated the feasibility of acquiring fundus photographs with portable devices and cell phones.[22,23,24] Although using SDOCT may still be relatively unfeasible in most screening settings, our approach demonstrates that a deep learning algorithm can closely replicate general SDOCT average RNFL thickness measurements from optic disc photos and could be potentially implemented in low-cost screening settings using fundus photographs. Further investigations should evaluate the feasibility and accuracy of such approaches.

As another important advantage of the quantitative approach for assessing optic disc photos presented in this work is the potential for assessing changes over time in settings where SDOCT is not available. A qualitative yes/no assessment of disc photos as performed previously does not generally allow assessment of changes over time, notably in those already classified as glaucomatous at baseline. By providing a continuous output, our approach could potentially be used to extract progression information from optic disc photographs that could be used for monitoring glaucomatous damage. However, validation of such an approach will require longitudinal investigations. 

The activation heatmaps showed that the locations in the optic disc photos that were most important for the deep learning algorithm corresponded very closely to the optic disc and adjacent RNFL, as seen in Figure 4. Retinal blood vessels or areas further from the optic disc had much smaller activations. This provides further confirmation that the algorithm is indeed identifying the area of the photo that is important for diagnosing glaucoma. The approach presented here may allow future investigations of features of optic discs that present the greatest challenges for recognition of signs of the disease, increasing awareness for their significance and opening opportunities for better training of clinicians on how to recognize them in clinical practice.

Our study has limitations. It should be noted that although the deep learning algorithm performed well in identifying glaucomatous damage, approximately 30$\%$ of the variance of SDOCT measurements remained unexplained. Many factors can explain this, such as variability of optic disc appearances and SDOCT measurements, as well as differences in photo quality. In fact, it would be surprising to find perfect correlations between photo predictions and SDOCT, considering that SDOCT obtains precise measurements of tissue thickness at a micrometer scale. We have also not performed a qualitative assessment of the optic disc photographs for quality. It is possible that removing poor quality photographs would result in even better performance of the algorithm. In fact, Figure 6 shows some random cases that were misclassified by the deep learning algorithm and it is possible to see that some of them had relatively low quality photographs. Furthermore, the algorithm was trained to replicate average thickness measurements from SDOCT rather than segmental SDOCT loss. It is likely that more sophisticated approaches could be created by having different SDOCT measurements as target values, including sectoral measurements or measurements from other areas of the optic disc or macula. In fact, in one of the examples shown in Figure 6 (second row, first photo from the left), the deep learning algorithm classification (abnormal, with probability of 72$\%$) disagreed with the SDOCT average RNFL classification (normal), but subjective analysis of the disc photo actually shows inferior localized rim thinning. Further training of the deep learning network with bigger datasets is likely to improve its performance even further.

In conclusion, we introduced a novel deep learning approach to assess optic disc photographs and provide quantitative information about the amount of neural damage. By analyzing disc photos, the deep learning algorithm was trained to closely replicate measurements obtained from average SDOCT RNFL thickness. The approach presented in this work could potentially be used to diagnose and stage glaucomatous damage from optic disc photographs.

\section{References}

[1] R. N. Weinreb, T. Aung, and F. A. Medeiros. The pathophysiology   and treatment of glaucoma: a review. JAMA,
311(18):1901-11, 2014.

[2] D. F. Garway-Heath, D. P. Crabb, C. Bunce, G. Lascaratos, F. Amalfitano, N. Anand, A. Azuara-Blanco, R. R.
Bourne, D. C. Broadway, I. A. Cunliffe, J. P. Diamond, S. G. Fraser, T. A. Ho, K. R. Martin, A. I. McNaught,
A. Negi, K. Patel, R. A. Russell, A. Shah, P. G. Spry, K. Suzuki, E. T. White, R. P. Wormald, W. Xing, and T. G. Zeyen. Latanoprost for open-angle glaucoma (ukgts): a randomised, multicentre, placebo-controlled trial. Lancet, 385(9975):1295-304, 2015.

[3] J. M. Tielsch, J. Katz, H. A. Quigley, N. R. Miller, and A. Sommer. Intraobserver and interobserver agreement in
measurement of optic disc characteristics. Ophthalmology, 95(3):350-6, 1988.

[4] R. Varma, W. C. Steinmann, and I. U. Scott. Expert agreement in evaluating the optic disc for glaucoma.
Ophthalmology, 99(2):215-21, 1992.

[5] H. D. Jampel, D. Friedman, H. Quigley, S. Vitale, R. Miller, F. Knezevich, and Y. Ding. Agreement among
glaucoma specialists in assessing progressive disc changes from photographs in open-angle glaucoma patients.
Am J Ophthalmol, 147(1):39-44 e1, 2009.

[6] H. H. Chan, D. N. Ong, Y. X. Kong, E. C. O'Neill, S. S. Pandav, M. A. Coote, and J. G. Crowston. Glaucomatous optic neuropathy evaluation (gone) project: the effect of monoscopic versus stereoscopic viewing conditions on optic nerve evaluation. Am J Ophthalmol, 157(5):936-44, 2014.

[7] Z. Li, Y. He, S. Keel, W. Meng, R. T. Chang, and M. He. Efficacy of a deep learning system for detecting
glaucomatous optic neuropathy based on color fundus photographs. Ophthalmology, 125(8):1199-1206, 2018.

[8] D. S. W. Ting, C. Y. Cheung, G. Lim, G. S. W. Tan, N. D. Quang, A. Gan, H. Hamzah, R. Garcia-Franco, I. Y.
San Yeo, S. Y. Lee, E. Y. M. Wong, C. Sabanayagam, M. Baskaran, F. Ibrahim, N. C. Tan, E. A. Finkelstein, E.  L. Lamoureux, I. Y. Wong, N. M. Bressler, S. Sivaprasad, R. Varma, J. B. Jonas, M. G. He, C. Y. Cheng, G. C. M. Cheung, T. Aung, W. Hsu, M. L. Lee, and T. Y. Wong. Development and validation of a deep learning system for diabetic retinopathy and related eye diseases using retinal images from multiethnic populations with diabetes. JAMA, 318(22):2211-2223, 2017.

[9] H. Takahashi, H. Tampo, Y. Arai, Y. Inoue, and H. Kawashima. Applying artificial intelligence to disease staging: Deep learning for improved staging of diabetic retinopathy. PLoS One, 12(6):e0179790, 2017.

[10] A. J. Tatham and F. A. Medeiros. Detecting structural progression in glaucoma with optical coherence tomography. Ophthalmology, 124(12S):S57-S65, 2017.

[11] C. K. Leung, C. Y. Cheung, R. N. Weinreb, Q. Qiu, S. Liu, H. Li, G. Xu, N. Fan, L. Huang, C. P. Pang, and D. S. Lam. Retinal nerve fiber layer imaging with spectral-domain optical coherence tomography: a variability and
diagnostic performance study. Ophthalmology, 116(7):1257-63, 1263 e1-2, 2009.

[12] Z. M. Dong, G. Wollstein, and J. S. Schuman. Clinical utility of optical coherence tomography in glaucoma.
Invest Ophthalmol Vis Sci, 57(9):OCT556-67, 2016.

[13] T. M. Kuang, C. Zhang, L. M. Zangwill, R. N. Weinreb, and F. A. Medeiros. Estimating lead time gained by
optical coherence tomography in detecting glaucoma before development of visual field defects. Ophthalmology, 122(10):2002-9, 2015.

[14] Kaiming He, Xiangyu Zhang, Shaoqing Ren, and Jian Sun. Deep residual learning for image recognition. CoRR, abs/1512.03385, 2015.

[15] J. Deng, W. Dong, R. Socher, L.-J. Li, K. Li, and L. Fei-Fei. ImageNet: A Large-Scale Hierarchical Image
Database. In CVPR09, 2009.

[16] Ramprasaath R. Selvaraju, Abhishek Das, Ramakrishna Vedantam, Michael Cogswell, Devi Parikh, and Dhruv
Batra. Grad-cam: Why did you say that? visual explanations from deep networks via gradient-basedlocalization.
CoRR, abs/1610.02391, 2016.

[17] F. A. Medeiros, P. A. Sample, L. M. Zangwill, J. M. Liebmann, C. A. Girkin, and R. N. Weinreb. A statistical
approach to the evaluation of covariate effects on the receiver operating characteristic curves of diagnostic tests in glaucoma. Invest Ophthalmol Vis Sci, 47(6):2520-7, 2006.

[18] H. L. Rao, U. K. Addepalli, S. Chaudhary, T. Kumbar, S. Senthil, N. S. Choudhari, and C. S. Garudadri. Ability of different scanning protocols of spectral domain optical coherence tomography to diagnose preperimetric glaucoma. Invest Ophthalmol Vis Sci, 54(12):7252-7, 2013.

[19] R. Lisboa, Jr. Paranhos, A., R. N. Weinreb, L. M. Zangwill, M. T. Leite, and F. A. Medeiros. Comparison of
different spectral domain oct scanning protocols for diagnosing preperimetric glaucoma. Invest Ophthalmol Vis Sci, 54(5):3417-25, 2013.

[20] N. B. Patel, J. L. Wheat, A. Rodriguez, V. Tran, and R. S. Harwerth. Agreement between retinal nerve fiber layer measures from spectralis and cirrus spectral domain oct. Optom Vis Sci, 89(5):E652-66, 2012.

[21] M. T. Leite, H. L. Rao, R. N. Weinreb, L. M. Zangwill, C. Bowd, P. A. Sample, A. Tafreshi, and F. A. Medeiros. Agreement among spectral-domain optical coherence tomography instruments for assessing retinal nerve fiber layer thickness. Am J Ophthalmol, 151(1):85-92 e1, 2011.

[22] M. P. Shanmugam, D. K. Mishra, R. Madhukumar, R. Ramanjulu, S. Y. Reddy, and G. Rodrigues. Fundus imaging with a mobile phone: a review of techniques. Indian J Ophthalmol, 62(9):960-2, 2014.

[23] C. Lamirel, B. B. Bruce, D. W. Wright, N. J. Newman, and V. Biousse. Nonmydriatic digital ocular fundus
photography on the iphone 3g: the foto-ed study. Arch Ophthalmol, 130(7):939-40, 2012.

[24] S. Kumar, E. H. Wang, M. J. Pokabla, and R. J. Noecker. Teleophthalmology assessment of diabetic retinopathy fundus images: smartphone versus standard office computer workstation. Telemed J E Health, 18(2):158-62, 2012.

\end{document}